\documentclass{article}

\usepackage{microtype}
\usepackage{graphicx}
\usepackage{subfigure}
\usepackage{booktabs} % for professional tables

\usepackage{amsmath}
\usepackage{amsfonts}
\usepackage{theorem}
\usepackage{color}
\usepackage{multirow}
\usepackage{float}

\usepackage{hyperref}

\usepackage[accepted]{icml2020}

\theorembodyfont{\normalfont}

\newtheorem{definition}{Definition}
\newtheorem{proposition}{Proposition}
\newtheorem{remark}{Remark}

\newcommand{\R}{\mathbb{R}}

\icmltitlerunning{Nested Subspace Arrangement for Representation of Relational Data}

\begin{document}

\twocolumn[
%\icmltitle{Disk-ANChor Arrangement (DANCAR) for Directed Graph Representation\\ and its Visualization}
\icmltitle{Nested Subspace Arrangement for Representation of Relational Data}



\icmlsetsymbol{equal}{*}

\begin{icmlauthorlist}
\icmlauthor{Nozomi Hata}{equal,gd}
\icmlauthor{Shizuo Kaji}{equal,imi}
\icmlauthor{Akihiro Yoshida}{gd}
\icmlauthor{Katsuki Fujisawa}{imi}

\end{icmlauthorlist}

%\icmlaffiliation{to}{Department of Computation, University of Torontoland, Torontoland, Canada}
\icmlaffiliation{imi}{Institute of Mathematics for Industry, Kyushu University, Fukuoka, Japan}
\icmlaffiliation{gd}{Graduate School of Mathematics,  Kyushu University, Fukuoka, Japan}

\icmlcorrespondingauthor{Nozomi Hata}{n.hata@kyudai.jp}

\icmlkeywords{Machine Learning, ICML, representation learning, graph embedding}
\vskip 0.3in
]



\printAffiliationsAndNotice{\icmlEqualContribution} % otherwise use the standard text.
\begin{abstract}
%Graph is a ubiquitous mathematical object to represent 
%relational structure to model various entities.
%However, due to its discrete nature, it often requires non-trivial pre-processing before a graph %is analysed or optimised using existing techniques. Therefore,
%it is an important task to give a continuous (vector) representation of a graph. blah-blah...
%}
Studies on acquiring appropriate continuous representations of discrete objects, such as graphs and knowledge base data, have been conducted by many researchers in the field of machine learning.
In this study, we introduce Nested SubSpace (NSS) arrangement, a comprehensive framework for representation learning.
We show that existing embedding techniques can be regarded as special cases of the NSS arrangement.
Based on the concept of the NSS arrangement, we implement a Disk-ANChor ARrangement (DANCAR), a representation learning method specialized to reproducing general graphs.
Numerical experiments have shown that DANCAR has successfully embedded WordNet in $\R^{20}$ with an F1 score of 0.993 in the reconstruction task.
DANCAR is also suitable for visualization in understanding the characteristics of graphs.
\end{abstract}

\section{Introduction}
\begin{table*}[ht]
\caption{The categorization of representation learning with respect to its input and representation space.}\label{tab:categorize}
\begin{center}
  \begin{tabular}{  c || c | c | c } \hline

    Method & $\mathcal{V}$ & $\mathcal{X}$ & Structure  \\ \hline
    Most existing method &undirected graph &points in $\R^k$ &metric or (dis)similarlity \\
    TransE &multi relational data &points in $\R^k$ &metric, addition \\
    Poincar\'{e} Embedding & hierarchical data &points in the Poincar\'{e} disk & metric \\
    Disk Embedding &directed acyclic graph &disks in a metric space &metric, inclusion \\ \hline
    {\bf DANCAR (Proposed)} &directed graph &anchored disks in a metric space &metric, inclusion\\ \hline

  \end{tabular}
\end{center}
\end{table*}

Studies on acquiring the appropriate continuous representation of discrete objects have been closely associated with machine learning. These studies aim to obtain the low-dimensional vector representations of the objects by preserving their characteristics. Recent algorithms of representation learning have broad applications, such as preprocessing in machine learning or visualization. As discrete objects, graphs, knowledge bases, and social networks are the primary research targets in these fields.

By representation, we mean the following:
let $\mathcal{V}$ be a set of objects with a certain discrete structure. An \emph{embedding} or a \emph{representation} of
$\mathcal{V}$ is a mapping $\Psi : \mathcal{V}\to \mathcal{X}$,
where $\mathcal{X}$ is a space parametrized by real vectors.
Through $\Psi$, the set $\mathcal{V}$ can be equipped with
various structures of $\mathcal{X}$ such as addition, multiplication, differentiation, metric, and topology so that 
various operations, analysis, optimization techniques become available to deal with the elements of $\mathcal{V}$.
Representation has served as a fundamental building block for
various algorithms such as 
classification, clustering, information retrieval, link prediction, and visualization. 

Below, we review some existing works according to the type of $\mathcal{V}$ and $\mathcal{X}$, categorized in Table~\ref{tab:categorize}.
The most classical and fundamental case is when $\mathcal{V}$ 
is the set of (non-directed, simple) graphs,
and $\mathcal{X}$ is point clouds in a Euclidean space $\R^k$, where
vertices of a graph are mapped as points in $\R^k$.
Existing works in this direction include 
matrix factorization models~\cite{chung97,cao2015grarep,ou2016asymmetric,singh2008relational,cox2000multidimensional,balasubramanian2002isomap}, 
random walk based models~\cite{grover2016node2vec,perozzi2014deepwalk,dong2017metapath2vec,pan2016tri,yanardag2015deep}, 
and others~\cite{kipf2016variational,wang2016structural,chami2019hyperbolic,khasanova2017graph,li2015gated,duvenaud2015convolutional,monti2017geometric,wang2018graphgan}.

A common problem with these methods is the determination of the embedding dimension $k$. If it is too small, the embedding does not preserve structures of $\mathcal{V}$.
To achieve a high fidelity embedding, using point clouds in the 
Poincar\'{e} disk as the target space $\mathcal{X}$ was proposed~\cite{DBLP:journals/corr/NickelK17}. 
In particular, they succeeded in obtaining a low-dimensional representation of tree-like graphs. 

In search of representations of more general graphs, the
Disk Embedding~\cite{pmlr-v97-suzuki19a} was proposed to deal with directed acyclic graphs (DAGs).
Their idea is to use as $\mathcal{X}$ the set of disks (balls) in a metric space such as a Euclidean space, a spherical space, or a hyperbolic space.
This approach generalizes existing works for embedding DAGs such as the Order Embedding~\cite{vendrov2015order} and the Hyperbolic Entailment Cones ~\cite{ganea2018hyperbolic}.
The key idea of the Disk Embedding is to represent an edge of a graph by an inclusion relation between two disks, which leads to a successful embedding for DAGs.

Moreover, some embedding algorithms for hypergraphs have been recently proposed~\cite{feng2019hypergraph,tu2018structural,yang2019revisiting}.

Other than graphs, knowledge bases have also been intensively studied by many researchers. 
A knowledge base consists of various relationships among entities represented as triples (head entity, relation, tail entity), e.g., (Vienna, {\tt IsCapitalOf}, Austria).
They are essential resources for many applications such as question answering, content tagging, fact-checking, and knowledge inference. 
TransE~\cite{bordes2013translating} is the first translation-based method, which embeds entities and relations in a Euclidean space with the latter represented by the differences in the former. 
Many extended versions of TransE have been proposed, such as TransH~\cite{wang2014knowledge}, STransE~\cite{nguyen-etal-2016-stranse}, 
Riemannian TransE\cite{suzuki2018riemannian} and
TorusE~\cite{ebisu2018toruse}. 

In this study, we propose the Nested SubSpace arrangement (NSS arrangement), a comprehensive framework for representation learning (\S 2). In the NSS arrangement, a node is represented by a nested subspace of a metric space. Our NSS arrangement generalizes existing embedding techniques. 
As a special case of the NSS arrangement, we also propose the Disk-ANChor ARrangement (DANCAR), which is an embedding method for directed graphs possibly with cycles (\S 3).
The DANCAR maps a node to a pair of a disk and a point contained in the disk. A directed edge is considered to be present when a disk contains the point of another pair.
This containment relation is not symmetric nor transitive, which vests the DANCAR a great representational capacity.

We demonstrated the DANCAR with two numerical experiments. First, we visualized a part of the Twitter network, where both the cluster structure and the hierarchical structure of the graph were successfully captured (\S 4).
Second, we conducted the reconstruction task using the WordNet to see the representation capacity of the DANCAR (\S 5). 
The experiment showed that the DANCAR successfully represented the graph, where an F1 score of 0.993 for the edge reconstruction task
was achieved by the embedding in a 20-dimensional Euclidean space.

Our contributions in this study are summarized as follows:
\vspace{-0.5\baselineskip} \begin{itemize}\setlength{\itemsep}{-2pt}
    \item We propose the NSS arrangement, a general framework to
represent relational data in a continuous space.
    \item As a special case of the NSS arrangement, we propose the DANCAR to represent directed graphs. 
    \item We show that the DANCAR can be used to visualize a large-scale network to reveal cluster structure and hierarchical structure.
    \item We show that the DANCAR can be used to represent a directed graph accurately in terms of the edge reconstruction task. 
\end{itemize}

\section{Nested SubSpace Arrangement}

In this section, we introduce the NSS arrangement to represent discrete entities and their relationships in a continuous space. 
The NSS arrangement generalizes many existing methods for representation learning.

\begin{figure}[htb]
    \centering
    \includegraphics[width=8cm]{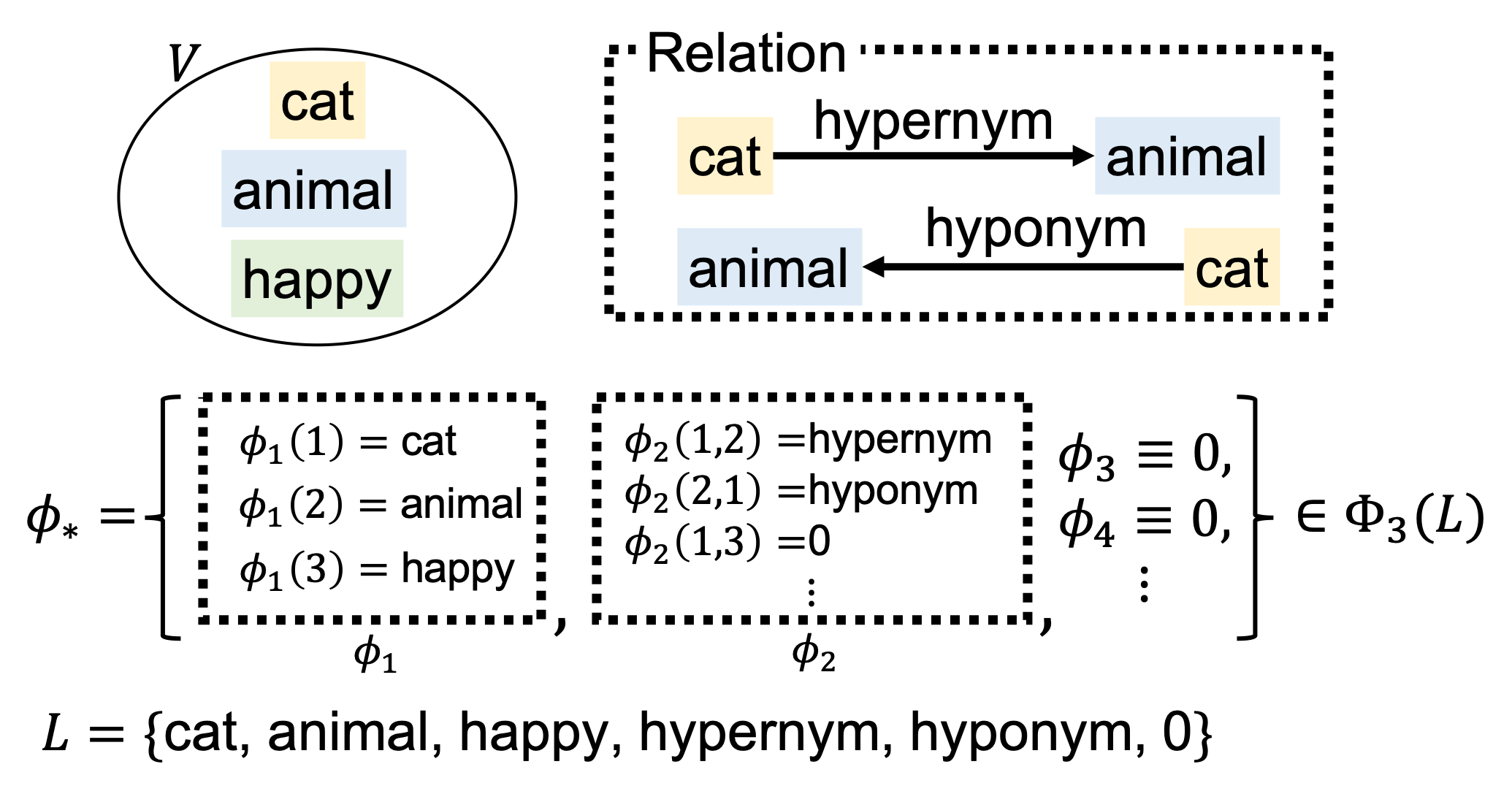}
    \caption{An example of the relational structure on a set $V$ of cardinality $3$.
    $\phi_1$ describes the name of each element in $V$, and $\phi_2$ describes the relation of each ordered pair of the elements.}
    \label{fig:relational_structure}
\end{figure}

Let $V$ and $L$ be discrete sets.
A \emph{relational structure} on $V$
with labels in $L$
is a sequence of maps
$\phi_* := \{ \phi_i: V^i\to L \mid i\in \mathbb{N} \}$. For example, a directed (non-simple) graph 
is expressed by $L=\{0,1,2,\ldots,\}$
and $\phi_i \equiv 0 \ (i\neq 2)$, where 
$\phi_2(u,v)$ is the number of directed edges from $u\in V$ to $v\in V$. 
Another example for multi-labeled network is shown in Figure~\ref{fig:relational_structure}.
Denote by $\Phi_m(L)$ the set of all the relational structures 
on a discrete set of cardinality $m$
with labels in $L$.

We call a triple $(V,L,\phi_*)$ \emph{relational data}.
Our purpose is to give a continuous representation of relational data.
We are particularly interested in 
\emph{binary} relational data in which $\phi_i \equiv 0 \ (i > 2)$.

\begin{definition}
Let $X$ be a metric space.
A sequence of spaces $A_1 \subset A_2 \subset \cdots \subset A_n \subset X$
is called a Nested SubSpace (NSS) with depth $n$ in $X$.
Denote by ${\mathcal S}_n(X)$ the set of all NSSs with depth $n$ in $X$. 
An ordered collection of NSSs is called an NSS arrangement.
\end{definition}

\begin{definition}
Fix $n,X,L$, and $V=\{v_1,\ldots,v_m\}$.
An \emph{embedding} of $(V,L,\phi_*)$
is a map $f:V \to {\mathcal S}_n(X)$.
A \emph{reconstruction} is a map
$g: ({\mathcal S}_n(X))^m\to \Phi_m(L)$.
\end{definition}
This means that each node in $V$ is represented by an NSS in $X$
and the relational data by an NSS arrangement in $X$.
The reconstruction map $g$ has to be defined in a rule-based manner according to
the type of relational structure.

The \emph{reconstruction task} for a fixed reconstruction $g$ is to find an embedding of $(V,L,\phi_*)$ such that $g(f(v_1),f(v_2),\ldots,f(v_m))$ is close to $\phi_*$.

\begin{figure}[h]
    \centering
    \includegraphics[width=8cm]{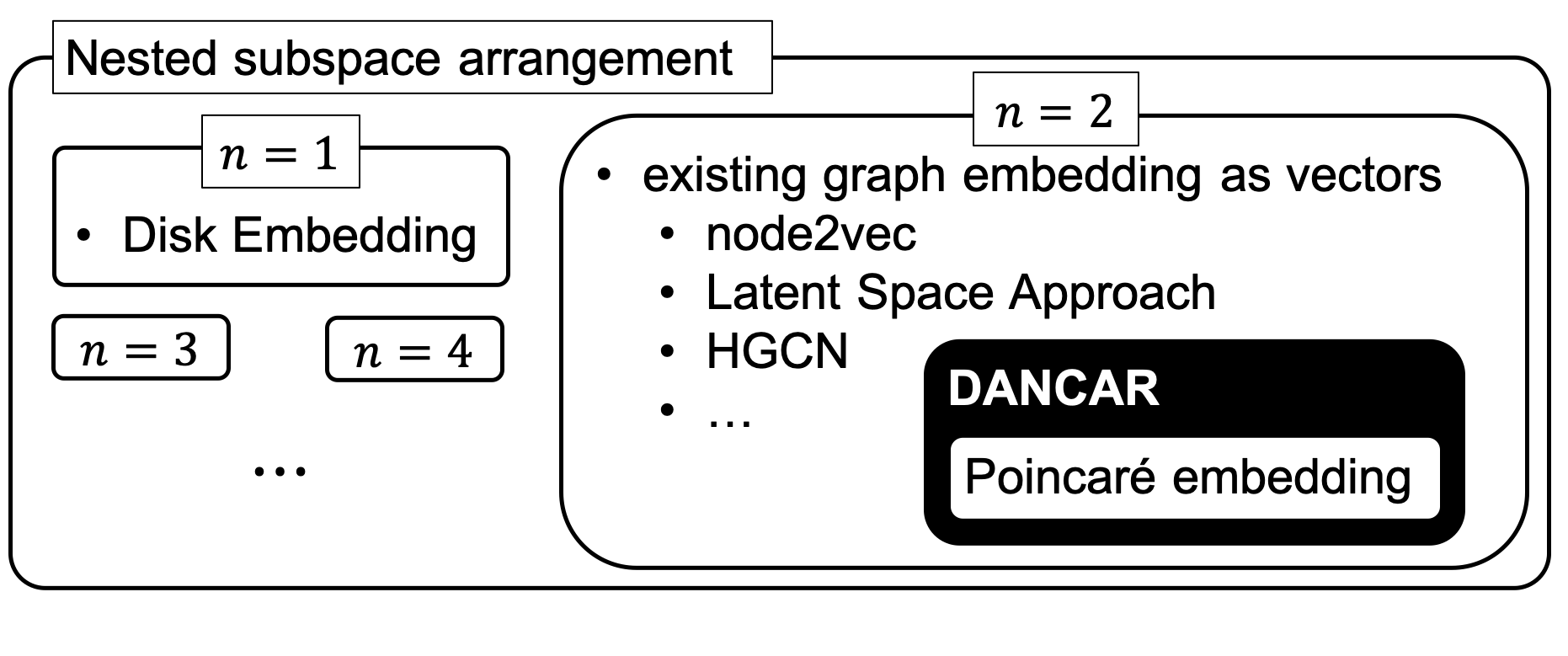}
    \caption{Hierarchy of NSS arrangement.}
    \label{fig:NSS_class}
\end{figure}

Two or more NSSs can be related in a various manner by containment of their members.
Our idea is to utilize this rich combinatorial structure among NSSs to represent binary (or possibly higher) relational data.
We illustrate the generality of the NSS arrangement by showing that the majority of existing methods can be regarded as special cases of the NSS arrangement (see Figure~\ref{fig:NSS_class}).

\begin{figure}[h]
    \centering
    \includegraphics[width=8cm]{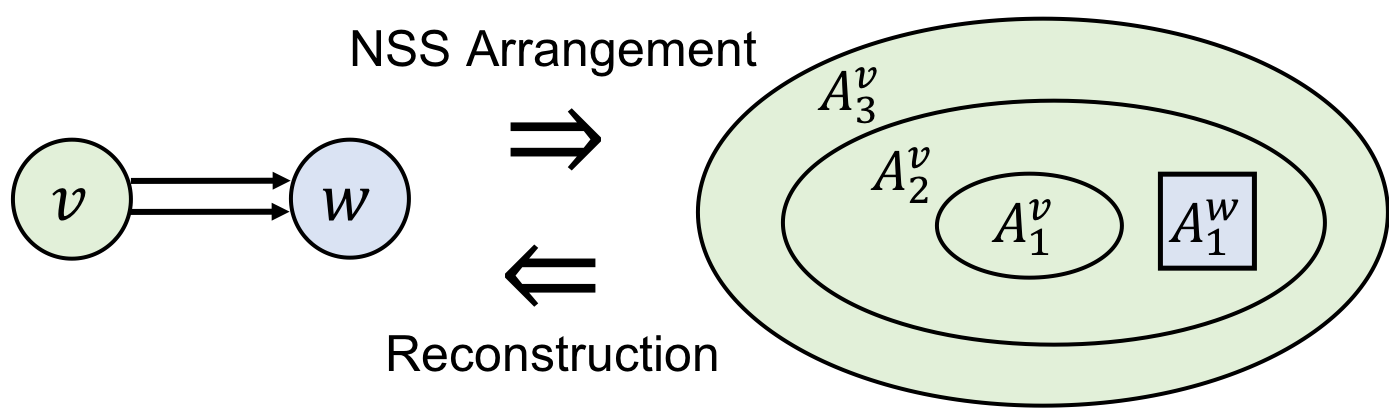}
    \caption{An example to represent multi-edge with NSS arrangement.}
    \label{fig:multiedge}
\end{figure}

Hereinafter, let $D(x,r)$ be the closed ball of radius $r>0$ centered at $x\in X$ in a metric space $(X,d)$.
\begin{enumerate}
\item[A.]{\bf{distance based}} (e.g., Poincar\'{e} Embedding\cite{DBLP:journals/corr/NickelK17})
A basic idea for graph embedding is to represent nodes as points, 
and edges are drawn between two points within a specified distance threshold $\epsilon>0$.
This is a special case of the NSS arrangement with
    \begin{itemize}
    \item $X$ is any metric space.
        \item $n = 2$. 
        \item {\it embedding}: A node $v \in V$ is mapped to an NSS $(\{x_v\} \subset D(x_v,\epsilon))$.
        \item {\it reconstruction}: an undirected edge $u$--$v$ is present when 
        $x_v\in D(x_u,\epsilon)$ (or equivalently, $x_u\in D(x_v,\epsilon)$).
    \end{itemize}
For example, the Poincar\'{e} embedding utilizes the metric structure on the Poincar\'{e} disk to represent undirected graphs.
\item[B.]{\bf{inner product based}} (e.g., \cite{chung97})
A slight variation of the previous example is to use inner product 
for reconstruction, where edges are drawn between points 
whose inner product is larger than a specified threshold $\tau>0$.
This is also a special case of the NSS arrangement with
\begin{itemize}
        \item $X=\R^k$ (or $S^{k-1}$).
        \item $n = 2$.
        \item {\it embedding}: A node $v \in V$ is mapped to an NSS 
        $(\{x_v\} \subset H(x_v,\tau)\cup \{x_v\})$,
        where $H(x_v,\tau) = \{y \in X \mid \langle x_v,y \rangle > \tau \}$.
        \item {\it reconstruction}: an undirected edge $u$--$v$ is present when $x_u \in H(x_v,\tau)$.
    \end{itemize}

\item[C.]{\bf{TransE}}
TransE~\cite{bordes2013translating} is the first translation-based model for representing multi-relational data, that is, $|L|>2$.
TransE maps a node $v$ to $x_v \in \R^k$ and a relation $l\in \{1,2,\ldots,s\}$ to $y_l\in \R^k$ to conform $x_u + y_l \approx x_v$
for a triple $(u,l,v)$.
This is another example of the NSS arrangement with a threshold $\epsilon>0$ with 
\begin{itemize}
    \item $X = \R^k$.
   \item $n = s+1$.
        \item {\it embedding}: A node $v \in V$ is mapped to an NSS 
        $(A^v_1=\{x_v+y_1\} \subset A^v_2=\{x_v+y_1,x_v+y_2\} \subset \cdots \subset
        A^v_{s+1}=\{x_v+y_1,x_v+y_2,\ldots,x_v+y_s\} \cup D(x_v,\epsilon))$.
        \item {\it reconstruction}: a relation $(u,l,v)$ is present when 
        $A^u_{l}\setminus A^u_{l-1} \subset D(x_v,\epsilon))$, where we regard $A_0^u=\emptyset$.
\end{itemize}

\item[D.]{\bf{Disk Embedding}}
Disk Embedding \cite{pmlr-v97-suzuki19a} can be regarded as an example of the NSS arrangement.
\begin{itemize}
    \item $n = 1$. 
    \item {\it embedding}: A node $v \in V$ is mapped to an NSS $D(x_v,r_v)$.
    \item {\it reconstruction}: A directed edge $(u,v)$ is reconstructed when $D(x_u, r_u) \subset D(x_v, r_v)$. 
\end{itemize}
Note that the Order Embedding~\cite{vendrov2015order} and the Hyperbolic Entailment Cones~\cite{ganea2018hyperbolic} are special cases of 
the Disk Embedding, thus are also examples of the NSS arrangement.

\item[E.]{\bf{Multi-graphs}}
The NSS arrangement can also represent a multi-graph (non-simple directed graph);
$k$-fold edges from $u$ to $v$ are represented by the relationship $A_1^u \subset A_{n-k+1}^v$ (see Figure~\ref{fig:multiedge}).
\end{enumerate}

\begin{figure}[h]
    \centering
    \includegraphics[width=8cm]{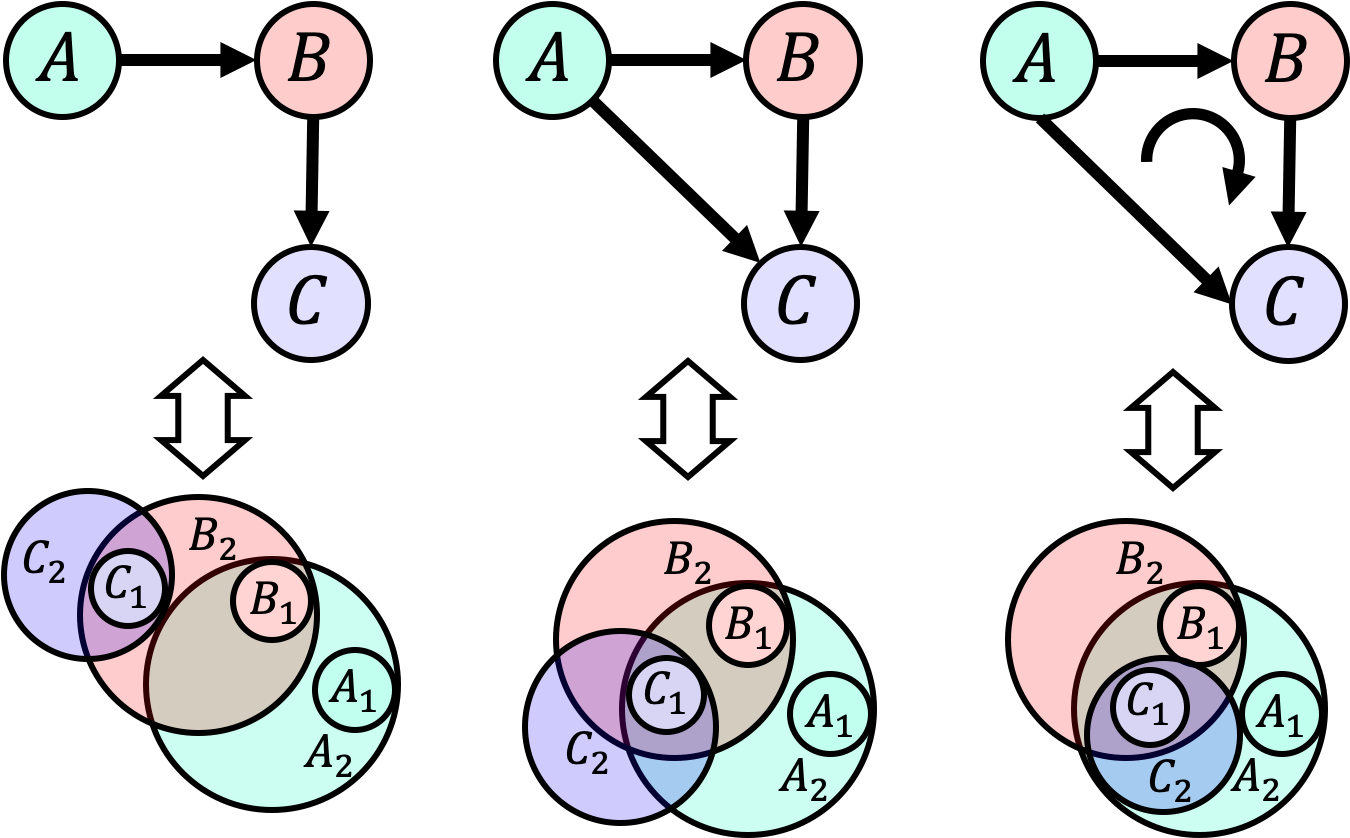}
    \caption{NSS can be used to capture higher relational data.
    Three types of relations among three objects above can be modelled by different containment relations of the corresponding NSSs.}
    \label{fig:higher_relational_data}
\end{figure}

\newpage
To close this section,
we explain the rich combinatorial structure of the NSS by a simple example.
First, note that Disk Embedding can only represent transitive relations:
$A \to B$ and $B\to C$ automatically imply $A \to C$.
In other words, Disk Embedding can represent only partially ordered sets.
On the other hand, the NSS can distinguish
the different relations among three objects depicted in Figure \ref{fig:higher_relational_data}
by assigning, for example, the following containment conditions:
\begin{align*}
\text{(left) } &B_1 \subset A_2, C_1 \subset B_2,  C_1 \not\subset A_2, C_2 \not\subset A_2\\
\text{(center) } &B_1 \subset A_2, C_1 \subset B_2, C_1 \subset A_2, C_2 \not\subset A_2\\
\text{(right) } &B_1 \subset A_2, C_1 \subset B_2, C_1 \subset A_2, C_2 \subset A_2. 
\end{align*}
The third one can be interpreted as a kind of directed ternary hyper-edge from $A$ to $(B,C)$.
The NSS can represent mathematical objects which are more general than partially ordered sets (see also Figure \ref{fig:comparison}).

\section{DANCAR : Disk-ANChor Arrangement}
We give an efficient implementation for
the embedding of directed graphs by
the NSS arrangement of depth $2$ in the Euclidean space $\R^k$, which we call 
the DANCAR.
\subsection{The DANCAR model}
\begin{figure}[htbp]
    \centering
    \includegraphics[width=8cm]{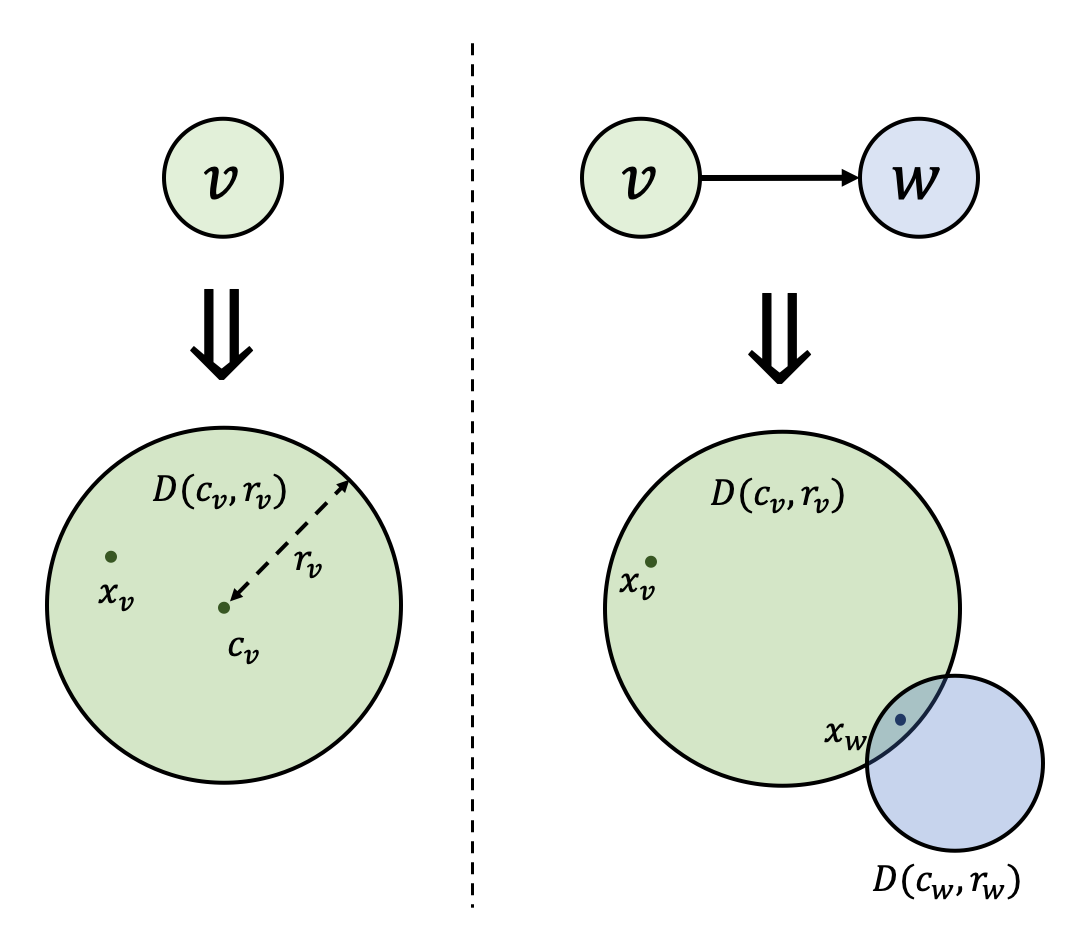}
    \caption{DANCAR represents a node by a pair of a disk and an anchor, and an edge by their membership relation.}
    \label{fig:howto_embedding_dancar}
\end{figure}

The DANCAR is a special case of the NSS arrangement:
\begin{itemize}\setlength{\itemsep}{-2pt}
    \item $X = \R^k$.
    \item $n = 2$. 
    \item {\it embedding}: $V\ni v \mapsto A_1^v=\{x_v\} \subset A_2^v = D(c_v, r_v) \subset X$, where $A_1^v=\{x_v\}$ is called the \emph{anchor} of the disk $ A_2^v = D(c_v, r_v)$.
    \item {\it reconstruction}: a directed edge $(v,w)$ is present when $x_w \in D(c_v, r_v)$ as illustrated in Figure~\ref{fig:howto_embedding_dancar}.
\end{itemize}

We optimize $x_v, c_v \in \R^k$ and $r_v > 0$ to find an embedding with a good reconstruction (see \S 3.3).
    
\begin{figure}[htbp]
    \centering
    \includegraphics[width=7cm]{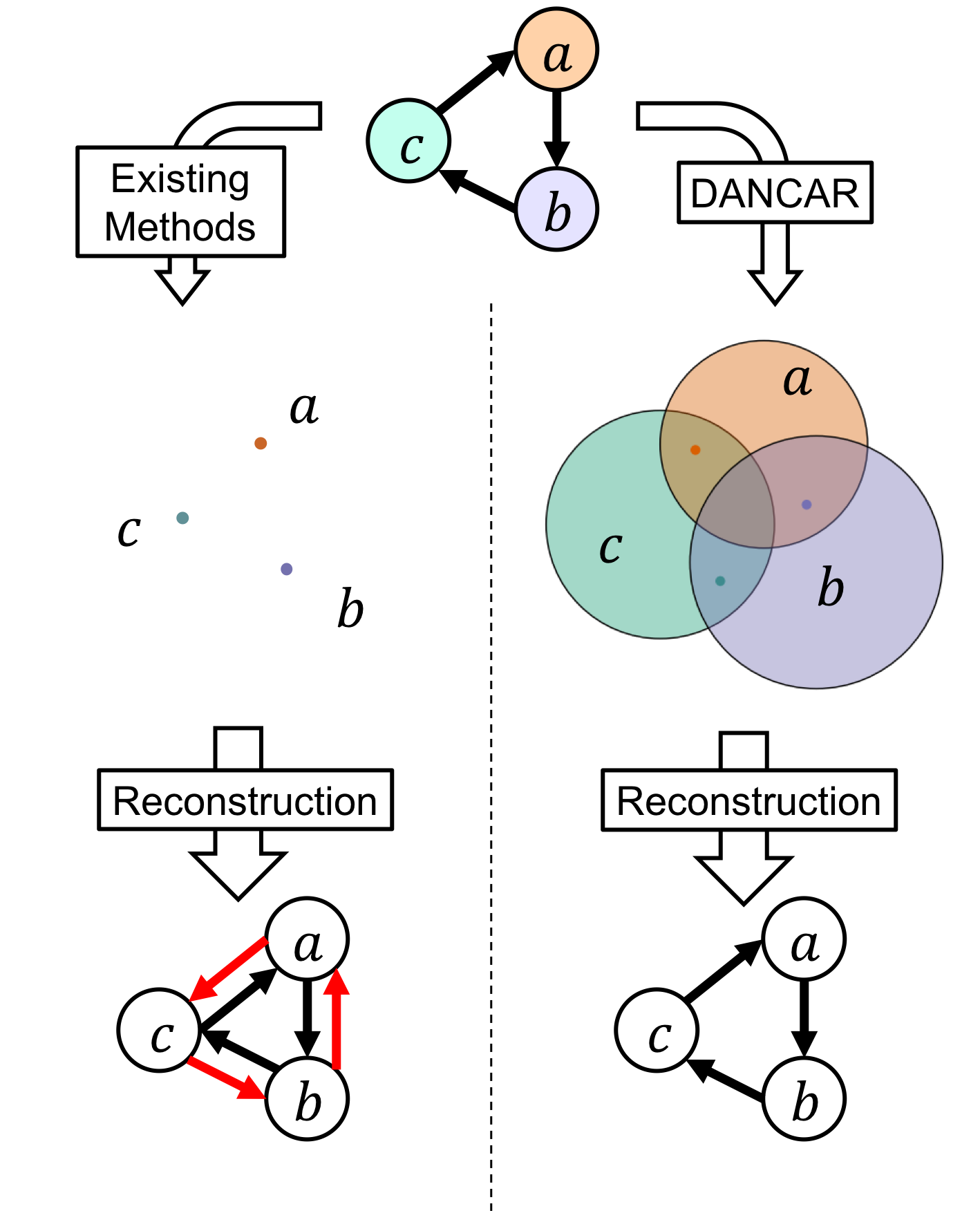}
    \caption{Directed cycles cannot be faithfully represented by a point cloud based embedding (left). Red arrows represent the edges which only exist in the reconstructed graph.
    DANCAR (right), on the other hand, can faithfully embed directed cycles.}
    \label{fig:comparison}
\end{figure}

Introduction of the anchor enables the DANCAR to 
represent non-symmetric and cyclic relations by containment (see Figure~\ref{fig:comparison}).

\subsection{Representational capacity of DANCAR}

At a glance, DANCAR is not so different from Disk Embedding, adding just anchor points.
However, this simple trick to add the anchor points provides a great representational power to DANCAR. DANCAR can represent directed cycles (Figure \ref{fig:comparison}) and 
 ``emulate'' the hyperbolic metric with the Euclidean metric (Proposition \ref{prop:poincare}).

First, it is easy to see the following proposition.
\begin{proposition}
Any directed tree can be embedded into $\mathbb{R}^2$ using the DANCAR.
\end{proposition}

\begin{figure}[h]
    \centering
    \includegraphics[width=4cm]{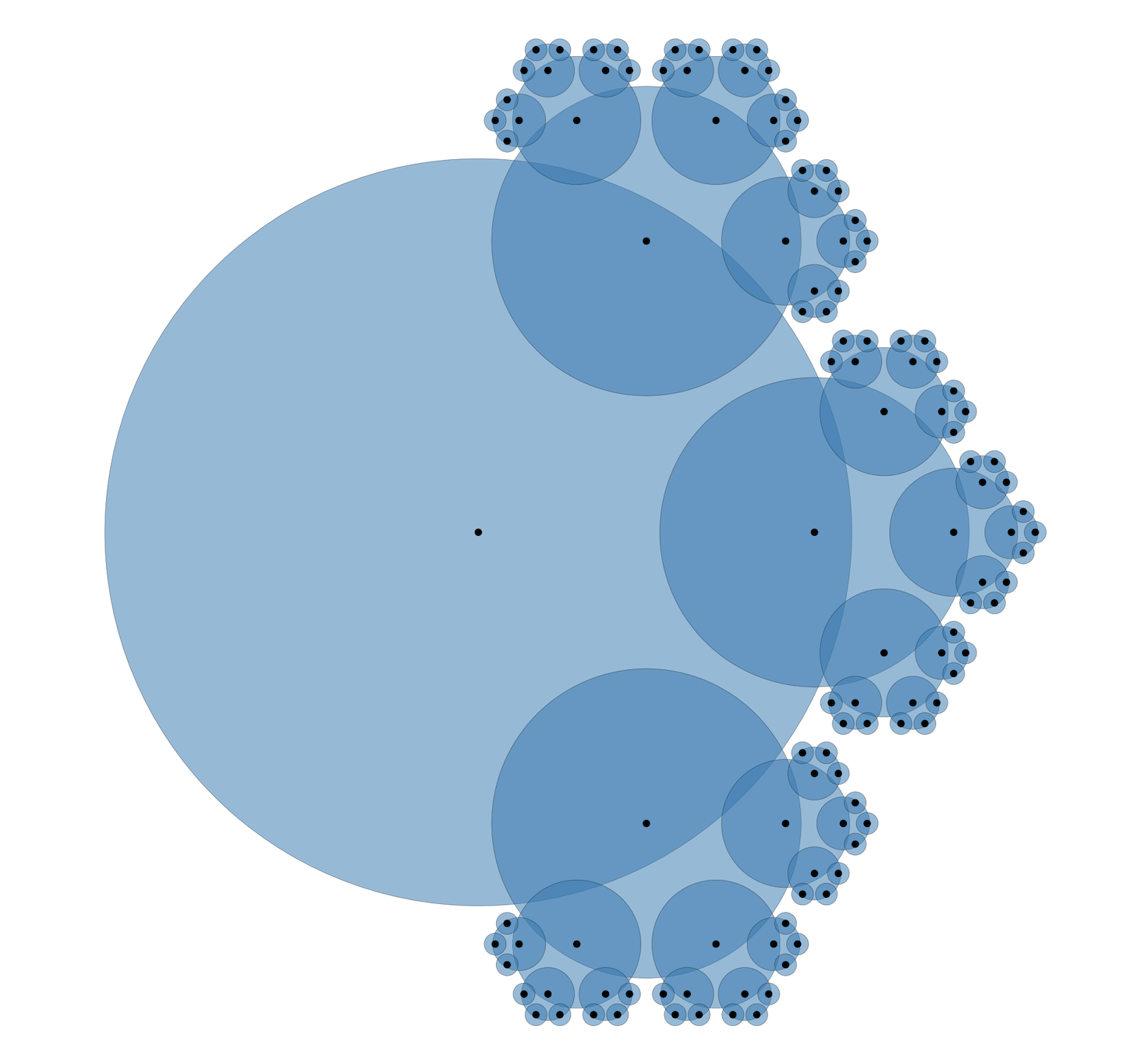}
    \caption{DANCAR embedding of the perfect ternary tree with $depth = 5$.}
    \label{fig:embedded_tree}
\end{figure}

Figure~\ref{fig:embedded_tree} 
graphically depicts how to embed a tree;
we can choose the radius of a node shrinking exponentially 
with respect to the distance from the root.
A concrete choice for the radius and the center position can be easily computed
(see Algorithm~\ref{alg:embed_tree}).

Less trivial is the fact that the DANCAR model generalizes the Poincar\'{e} embedding model.
In fact, since any tree can be faithfully embedded by the 2-dimensional Poincar\'e embedding, 
the above Proposition is a corollary of the following.
\begin{proposition}
\label{prop:poincare}
The Poincar\'{e} embedding is a special case of the DANCAR which satisfies 
\begin{enumerate}\setlength{\itemsep}{-2pt}
\item $c_v = x_v / (K_v+1)$,
\item $\|x_v\| < 1$,
\item $r_v = \sqrt{ \frac{K_v}{K_v+1} \left( 1-\frac{1}{K_v+1} \|x_v\|^2 \right) }$,
\end{enumerate}
where
\[
K_v := \frac{\cosh{r}-1}{2} (1-\|x_v\|^2).
\]
\end{proposition}

This follows from the fact that a sphere in a Poincar\'e disk is 
also a sphere in the Euclidean space but with a different radius and a center.
We provide a formal proof in Appendix.
Proposition~\ref{prop:poincare} indicates that our model can replicate the result of Poincar\'{e} embedding using the standard Euclidean norm.

\begin{figure}[htbp]
    \centering
    \includegraphics[width=8cm]{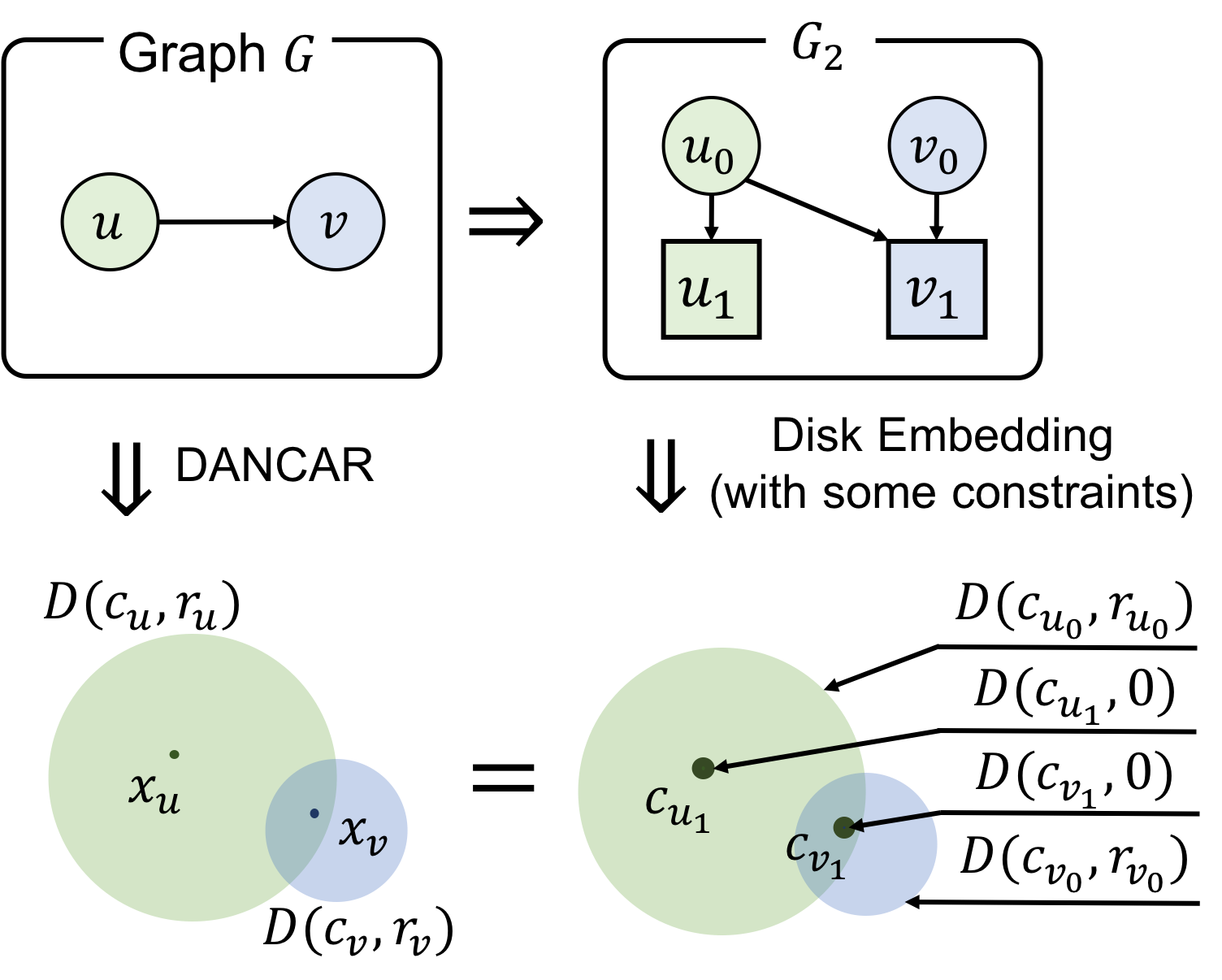}
    \caption{DANCAR as Disk Embedding of a transformed graph.}
    \label{fig:DE_DANCAR}
\end{figure}

\begin{remark}
We can view the DANCAR as the combination of a graph transformation and the Disk Embedding
(see Figure \ref{fig:DE_DANCAR}).
Given a directed graph $G=(V,E)$, let $G_2 = (V_2, E_2)$ be the directed bipartite graph defined by 
\begin{align*}
V_2 &:= \{u_i \mid  u \in V, i \in \{0,1\}\} \\
E_2 &:= \{(u_0, v_1) \in V_2^2 \mid (u,v) \in E \ {\text or}\  u = v\}.
\end{align*}
The DANCAR embedding $\{(x_v, D(c_v, r_v))\}_{v \in V}$ is identified with the Disk embedding of $G_2$,
where the radius for $v_1$ nodes are fixed to zero.
\end{remark}

\subsection{Construction of an embedding}
\begin{figure}[htbp]
    \centering
    \includegraphics[width=8cm]{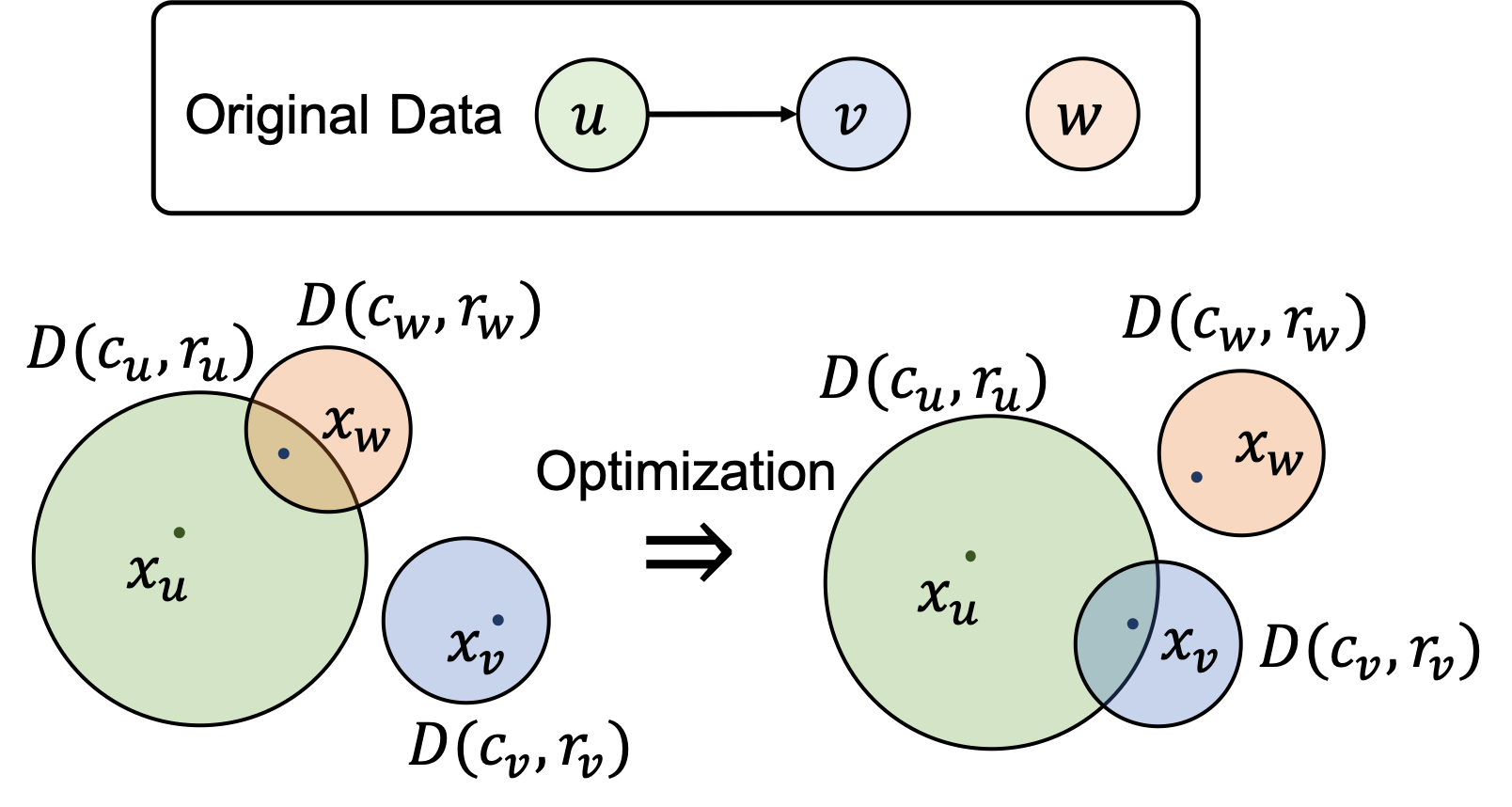}
    \caption{Illustration of an optimization process.}
    \label{fig:loss_function}
\end{figure}

Let $(V,E)$ be a directed graph.
We formulate the problem of finding a good embedding
of $(V,E)$ by the DANCAR as an optimization problem (see Figure~\ref{fig:loss_function}).
We introduce three loss functions with a hyperparameter, $\mu$, called {\it margin}:
\begin{itemize}
    \item Positive Loss : If there is a directed edge $(v,w)$, the anchor of the tail node $w$ should be included by the disk of the head node $v$.
    \begin{equation}
        L_{\rm pos} := \dfrac{1}{|E|}\sum_{(v,w) \in E} {\rm ReLU}(d(c_v,x_w) -  r_v +\mu).
    \end{equation}
    \item Negative Loss : If there is no directed edge $(v,w)$, the anchor of the tail node $w$ should not be included by the disk of the head node $v$.
    \begin{equation}
        L_{\rm neg} := \dfrac{1}{|E^c|}\sum_{(v,w) \in E^c} {\rm ReLU}(r_v - d(c_v,x_w) +\mu),
    \end{equation}
    where $E^c=\{(v,w)\in V\times V\mid (v,w)\not\in E, v\neq w\}$.
    When sampling from $E^c$ is computational intractable, 
    we approximate $E^c$ by $\{(v,w)\in V\times V\mid v\neq w\}$. 
    \item Anchor Loss : the anchor should be contained in the disk. This can be regarded as a regularization.
    \begin{equation}
        L_{\rm anc} := \dfrac{1}{|V|}\sum_{v \in V}{\rm ReLU}(d(c_v,x_v) - r_v + \mu).
    \end{equation}
\end{itemize}

The total loss function of the DANCAR can be written as the weighted sum of the above loss functions:
\begin{align}\label{eq:total_loss}
\begin{split}
&L_{\rm DANCAR}(\{c_v\}_{v \in V},\{r_v\}_{v \in V},\{x_v\}_{v \in V})\\ := &L_{\rm pos} + \lambda_{\rm neg} L_{\rm neg} + \lambda_{\rm anc} L_{\rm anc}, 
\end{split}
\end{align}
where $\lambda_{\rm neg}\ge 0$ and  $\lambda_{\rm anc}\ge 0$ are hyperparameters.

An embedding is obtained by optimizing the total loss function by a stochastic gradient descent.

\subsection*{Implementation}
All experiments were implemented in Chainer 7.4.0. 
Source code is publicly available at \url{https://github.com/KyushuUniversityMathematics/DANCAR}.

\section{Experiment: Visualization}
In this section, we show the potential of the DANCAR in visualizing graphs. 

\begin{figure}[ht]
    \includegraphics[width=8cm]{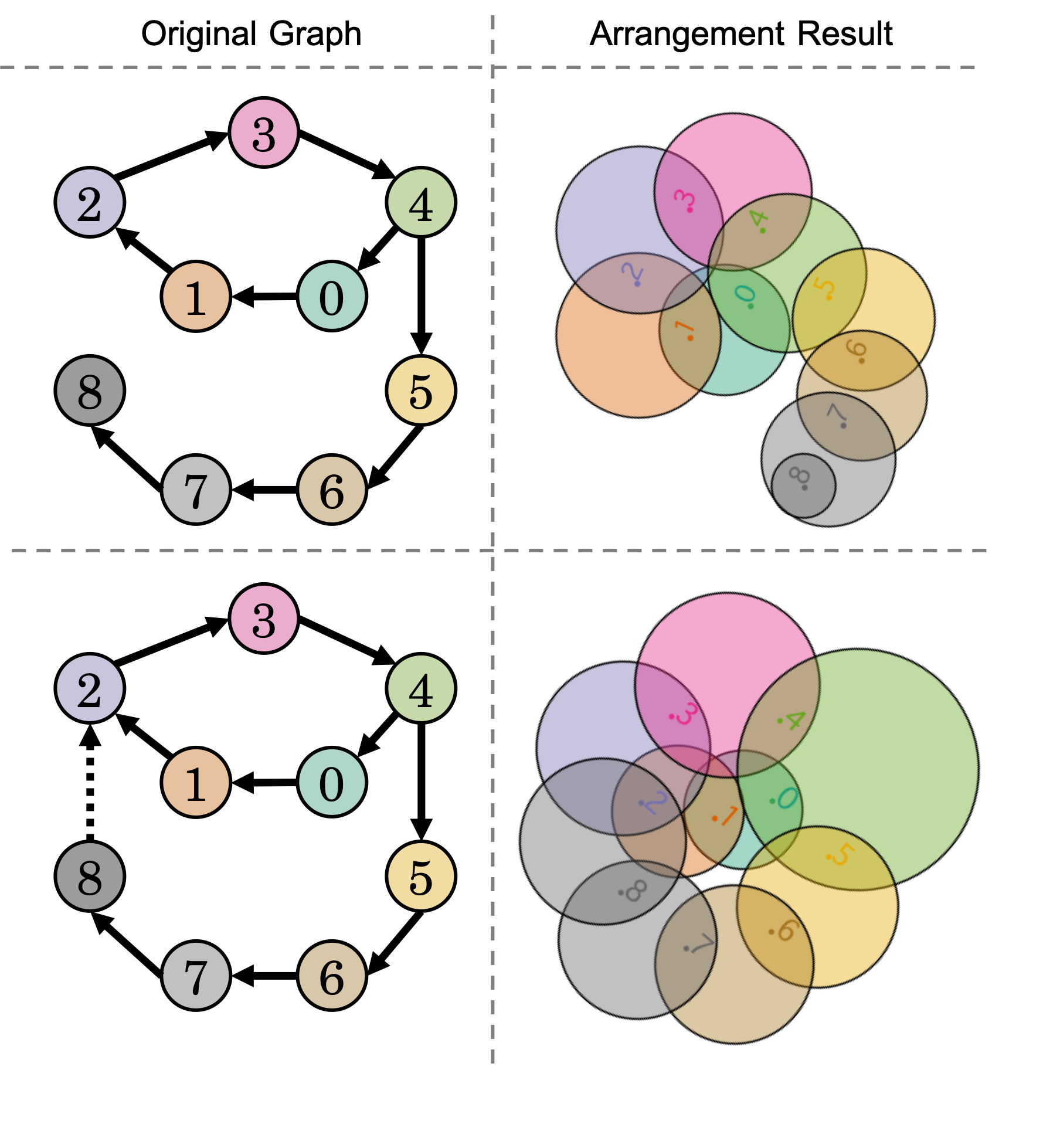}
    \caption{
    Embeddings by the DANCAR are sensitive to the change in the topology of the graph.}
    \label{fig:DANCAR_sensitivity}
\end{figure}

Figure~\ref{fig:DANCAR_sensitivity} 
illustrates with a toy example that
the DANCAR embedding is sensitive to the change in the topology of the graph.
The embedding faithfully captures the difference of two graphs, which is the existence of the edge $(8,2)$, illustrated as a broken line.

\begin{figure*}[htbp]
    \centering
    \includegraphics[width=17cm]{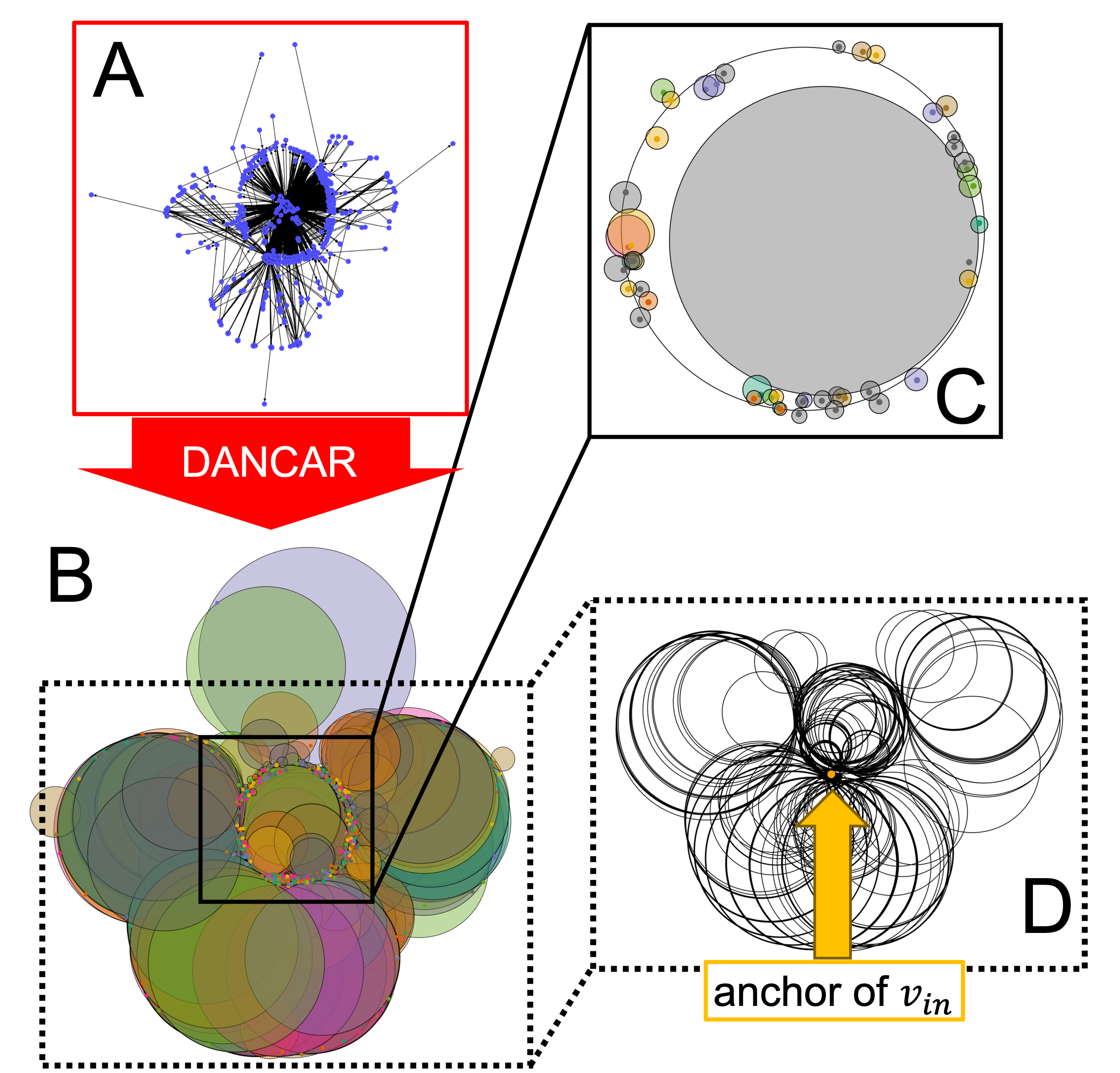}
    \caption{
    The visualization of a part of the Twitter network by the DANCAR.
    (A) a part of the Twitter network,
    (B) its DANCAR embedding in $\R^2$,
    (C) Neighbor of the node with the highest out-degree,
    (D) Neighbor of the node with the highest in-degree.}
    \label{fig:twitter_1000}
\end{figure*}

For a practical application,
we applied the DANCAR to a subgraph of the Twitter network\footnote{\url{https://snap.stanford.edu/data/twitter-2010.html}}. 
Each directed edge $(u,v)$ represents that  account $v$ follows account $u$.
We randomly picked 1,000 accounts, keeping the weak connectivity of the graph. 
The graph has 3,188 edges with the maximum in-degree 405 and the maximum out-degree 47.
The result of a DANCAR embedding of the graph into the two-dimensional Euclidean space
is shown in Figure~\ref{fig:twitter_1000}. 
Both clusters and hierarchies of the graph
can be observed through the visualization. 
The size of each disk roughly corresponds to  the out-degree of the node.
In fact, Spearman's rank correlation coefficient between the radius and out-degree is 0.628.

Figure~\ref{fig:twitter_1000} (C) shows 
the account $ v_ {out} $ with the highest out-degree and its successors (followers).
The disk of $ v_ {out} $ is depicted by the large black circle as $v_{out}$ is followed by a large number of other accounts.
On the other hand, we see that most of the accounts that follow $v_{out}$ have small radii (except for the gray one), reflecting the fact that 
they are not followed by many accounts; in fact, 
they are followed by at most one account in the original network.

Figure~\ref{fig:twitter_1000} (D) focuses on
the account $ v_ {in} $ with the highest in-degree and its predecessors.
The anchor and the disk (of tiny radius)
of $v_{in}$ are depicted by the yellow dot.
We observe that there are disks of various sizes around $v_{in}$.
The observation reflects the fact that 
$ v_ {in} $ follows both popular and non-popular accounts.

\section{Experiment: Reconstruction and Link prediction}
\begin{table*}[htbp]
\caption{The precision and the F1 score of the reconstruction (100\% training)
and the link prediction (50\% training) tasks. 
For the link prediction task, the evaluation scores were computed for the entire edges.
}\label{tab:numerical_experiment}
\begin{center}
  \begin{tabular}{ c || c || c | c  || c | c  || c | c  || c | c ||} \hline

    \multirow{6}{*}{\rotatebox[origin=c]{90}{WordNet}}&\multirow{3}{*}{method} &\multicolumn{4}{c||}{100\% training} &\multicolumn{4}{c||}{50\% training} \\ \cline{3-10}
    & &\multicolumn{2}{c||}{$10$dim} &\multicolumn{2}{c||}{$20$dim} &\multicolumn{2}{c||}{$10$dim} & \multicolumn{2}{c||}{$20$dim} \\ \cline{3-10}
    &  &F1 score & mAP  &F1 score & mAP &F1 score & mAP &F1 score & mAP\\ \cline{2-10}
    &{\bf DANCAR (Proposed)} 
    &0.982 &-  & 0.993 &- & 0.787 &- & 0.709 &- \\
    &{\bf Poincar\'{e} Embedding} & - & 0.635 & - & 0.654 &- & 0.675 &- &0.675\\
    &{\bf Disk Embedding} &0.057 & - &0.052 & - & 0.151 &- & 0.114 &-\\ \hline
  \end{tabular}
\end{center}
\end{table*}

\begin{figure}[h]
    \centering
    \includegraphics[width=7cm]{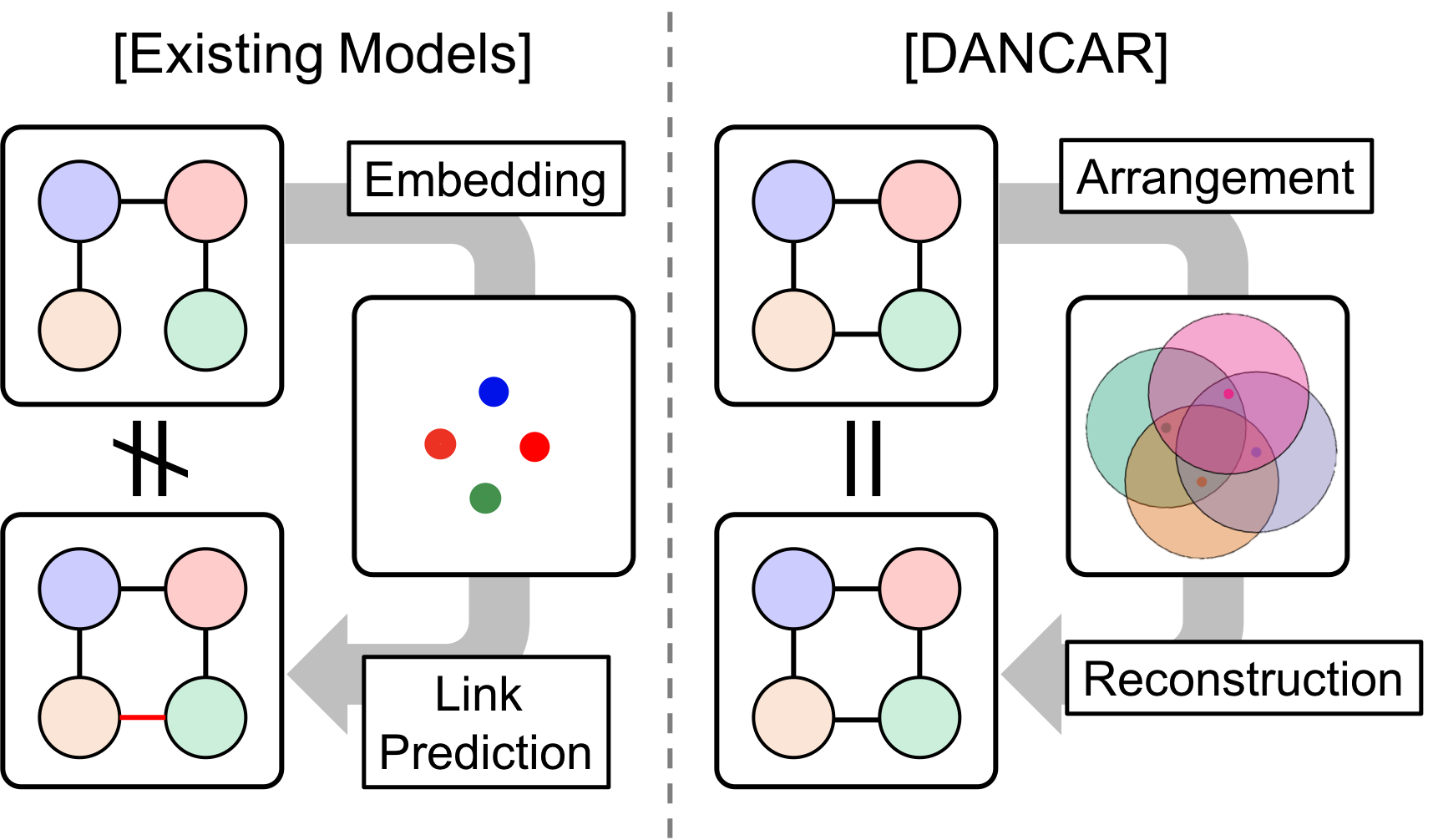}
    \caption{Non-existing edges in the training graph should be 
    reconstructed in the link prediction task, where as 
    they should not in the reconstruction task.}
    \label{fig:link_reproduction}
\end{figure}

In this section, we evaluate how well DANCAR can
represent directed graphs by
the reconstruction and the link prediction tasks.
The purpose of DANCAR is to faithfully capture the existence and the non-existence of edges.
Note that this is, in some sense, opposite to the link prediction task (Figure~\ref{fig:link_reproduction}). 
Nonetheless, we show that DANCAR performs well for both tasks by choosing appropriate 
embedding dimensions.

For the reconstruction experiment,
we compared the edge existence in the original graph and the reconstructed graph from the embedding.
For the link prediction experiment,
we computed the embedding using a half of edges randomly chosen from the original graph and
compared the edge existence in the original graph and the reconstructed graph from the embedding.

As a practical target graph\footnote{We observed that the subgraph of the Twitter network we used in the previous section was successfully embedded in the 10 dimensional Euclidean space perfectly faithfully (that is, with F1 score one).},
we used the largest weakly-connected component of a noun closure of the
{\bf WordNet}~\cite{miller1995wordnet}.
We removed the root and then took the transitive closure and obtained
 a DAG consisting of
 82,076 nodes and 660,846 directed edges.

The hyper-parameters for the DANCAR were chosen as follows.
The margin parameter $\mu$ was fixed to $0.01$. We tested with $\R^{10}$ and $\R^{20}$ as the embedding space.
We experimented with the hyper-parameters 
$8\le \lambda_{neg} \le 1000, \lambda_{anc}\in \{1,10\}$ and the best results were chosen. We observed that with higher embedding dimension, smaller 
$\lambda_{neg}$ performed better.
For optimization with a stochastic gradient descent, we used two different batch sizes 
$b_1=10,000$ for the positive loss and the vertex loss, 
and $b_2=100,000$ for the negative loss
to account for the sparsity of the graph.
We randomly selected the negative samples for each iteration.
We used the Adam~\cite{kingma2014adam} optimizer with parameters $\alpha=0.05, \beta_1 = 0.9$, and $\beta_2 = 0.999$.

Initialization of the parameters $c_v,r_v,x_v$ have been observed to be important, and we set:
\vspace{-0.5\baselineskip} \begin{itemize}\setlength{\itemsep}{-2pt}
    \item $c_v$ were sampled from the uniform distribution on $[-1,1]^k \ (k=10,20)$.
    \item $r_v=0.1$ for any $v\in V$.
    \item $x_v = c_v$ for any $v\in V$.
\end{itemize}\vspace{-0.5\baselineskip} 
This initial arrangement represents a graph with few edges.
Thus, in the beginning, the positive loss was dominant, and
the chance of the gradient vanishing problem 
was reduced.

As a comparison, we performed the same task with the Poincar\'{e} embedding and the Disk Embedding.
For the Poincar\'{e} embedding, we used the implementation made available by the original authors of \cite{DBLP:journals/corr/NickelK17}\footnote{https://github.com/facebookresearch/poincare-embeddings}.
We took 50 negative samplings per positive sample for the optimization of the Poincar\'{e} Embedding. For the Poincar\'{e} embedding, instead of choosing a single radius for all vertices for reconstruction, 
we used the mean average precision (mAP) for evaluation as was done in the original paper \cite{DBLP:journals/corr/NickelK17}.
Note that this is in a sense choosing an optimal radius for each vertex.
There is trade-off between precision and recall, and the F1 score is maximized when they agree.
Therefore, in most cases mAP is much higher than the F1 score computed for a choice of the radius.

For the Disk Embedding, we used our own implementation since our implementation is quite similar to the implementation of the Disk Embedding.
We used the same parameters and as the DANCAR except for the parameters for the anchors.

\begin{figure}[h]
    \centering
    \includegraphics[width=7cm]{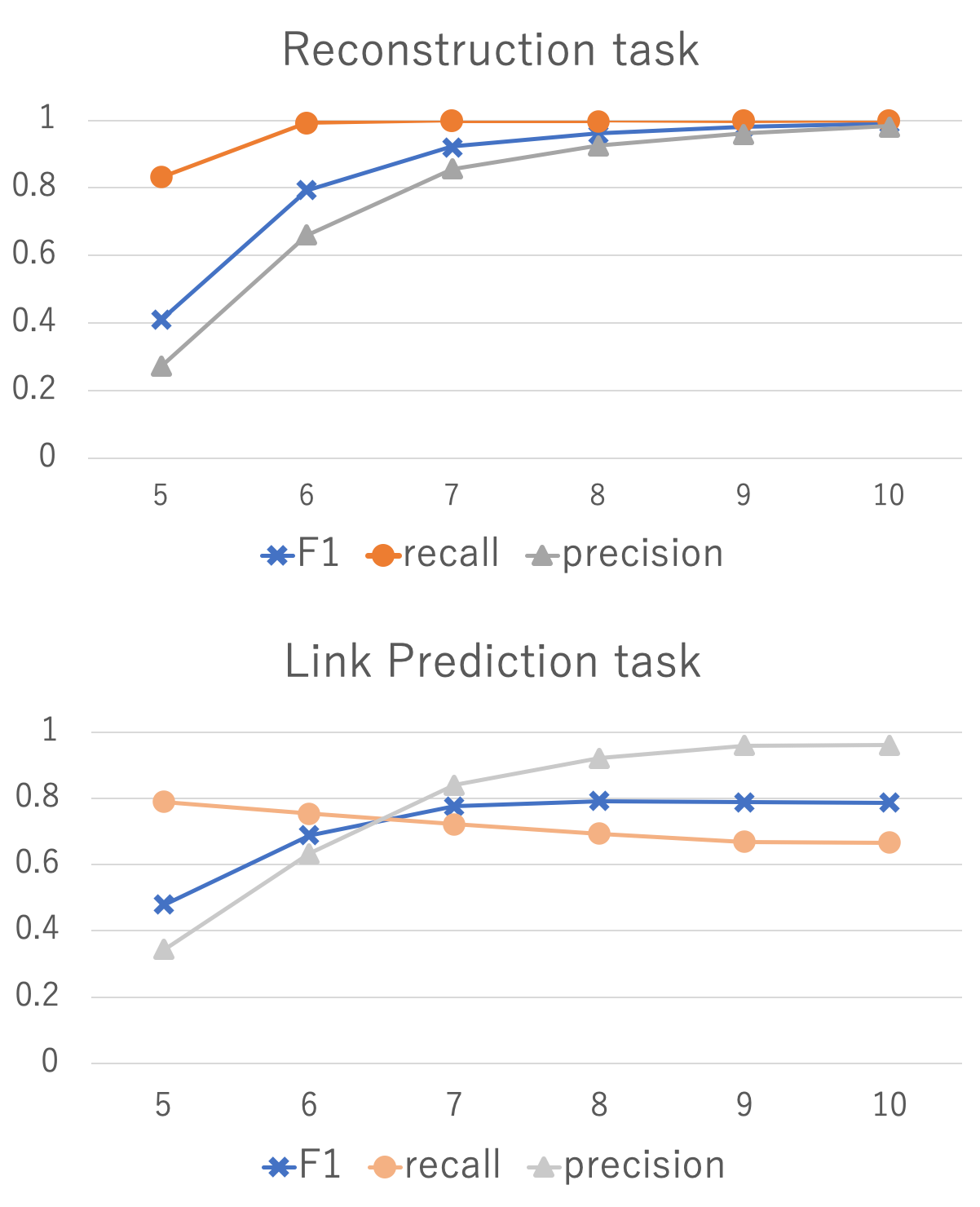}
    \caption{
    The performance of the reconstruction and the link prediction tasks
    with varying embedding dimensions.
    }
    \label{fig:dancar_reconstruct}
\end{figure}

The result is shown in Table~\ref{tab:numerical_experiment}.
We observed that our DANCAR performed
considerably better than the Poincar\'e embedding and the Disk Embedding.
We speculate that the absence of the root node in the graph has affected the performance of the Poincar\'e embedding and the Disk Embedding.
In contrast, our method does not depend on the existence of the root and 
was able to reconstruct the graph effectively.
It should be noted that in the link prediction task,
due to the high representational capacity of DANCAR,
the higher dimensional embeddings result in lower recall rate (see Figure~\ref{fig:dancar_reconstruct}).
When the embedding dimension is high enough, 
the DANCAR faithfully captures the (non-)existence of edges, and those edges 
which were not present in the training data were not reconstructed.
A similar phenomena should be observed for the Disk Embedding and 
the Poincar\'e embedding when we use a huge dimensional embedding space.

\section{Conclusion}
In this study, we introduced the Nested SubSpace (NSS) arrangement, which generalizes many of the existing methods for relational data representation.
As a special case of NSS arrangement, 
we provided a practical implementation 
of the Disk-ANChor ARrangement (DANCAR).

The visualization and large-scale embedding experiments highlighted the representation capacity of the DANCAR. We observed that the DANCAR captures the cluster and hierarchy structures simultaneously. 
The DANCAR outperformed existing methods in the reconstruction and the link prediction tasks of a large-scale DAG in a relatively low dimensional space.
The rich combinatorial structure of the DANCAR lead to an accurate representation of graphs.

Although in this paper we focused mainly on the DANCAR, 
our general framework of the NSS arrangement could be used for learning representation of 
various relational data.

\appendix
\begin{algorithm}[htbp]
   \caption{Tree embedding}
   \label{alg:embed_tree}
\begin{algorithmic}
   \STATE {\bfseries Input:} tree $G=(V,E)$ with the root $z \in V$
   \STATE {\bfseries Output:} the set of embedded disks $\{D(c_v, r_v)\}_{v \in V}$
   \STATE Let $n=\max_{v\in V}(\#\mathcal{N}(v))$, $\alpha = -\frac{(n-1)\pi}{2n}$, $p=\cos(\alpha)$, $q=\cos(2\alpha)$
   \STATE $t = \frac{\sqrt{(p+q)^2+4p} - p + q}{2(q + 1)}$, $k = \frac{1}{\sqrt{1+t^2}}$
   \STATE ${\boldsymbol c}_z = {\boldsymbol 0}, r_z = 1, \theta_z = 0$, $S = \{z\}$
   \WHILE{$S \neq \emptyset$}
        \STATE pop $u$ from $S$
        \STATE Let $\varphi = \alpha$
        \FOR{$v \in {\mathcal N}(u)$}
            \STATE $\theta_v = \theta_u + \varphi$
            \STATE $c_v = c_u + r_u k (\cos(\theta_v), \sin(\theta_v) )$
            \STATE $r_v = t r_u$
            \STATE push $v$ to $S$
            \STATE $\varphi = \varphi + \frac{\pi}{n}$
        \ENDFOR
    \ENDWHILE
\end{algorithmic}
\end{algorithm}
\subsection*{Appendix:Proof of Proposition 2}
Let $d(x,y)$ be the metric on the Poincar\'{e} ball, i.e., 
\[
d(x,y) := \textrm{arcosh} \left( 1 + 2\frac{\| x-y \| ^2}{\left( 1 - \|x\|^2\right) \left( 1 - \|y\|^2\right)} \right).
\]
Let $D_P(x,r)$ (respectively, $D_E(x,r)$) be the closed ball centered at $x$ and of radius $r>0$ with respect to the metric $d$ (respectively, the Euclidean metric).

For all $r>0$ and $a,x \in D_E(0,1)$, we have the following equivalence.
\begin{align*}
&x \in D_P(a,r) \\
\iff &d(x,a) \le r \\
\iff &\frac{\| x-a \| ^2}{\left( 1 - \|x\|^2\right) \left( 1 - \|a\|^2\right)} \le \frac{\cosh r - 1}{2} \\
\iff & \|x\|^2 - 2 \langle x,a \rangle + \|a\|^2 
\le K \left( 1 - \|x\|^2\right) \\
\iff & (K+1) \|x\|^2 - 2\langle x,a\rangle \le K - \|a\|^2 \\
\iff & (K+1) \left\| x - \frac{1}{K+1}a \right\|^2 - \frac{\|a\|^2}{K+1} \le K - \|a\|^2 \\
\iff & \left\| x - \frac{1}{K+1}a \right\|^2 \le \frac{K}{K+1}\left( 1 - \frac{1}{K+1}\|a\|^2 \right) \\
\iff &x \in D_E \left( \frac{1}{K+1}a, \sqrt{\frac{K}{K+1}\left( 1 - \frac{1}{K+1}\|a\|^2 \right)} \right)
\end{align*}
where 
\[
K := \frac{\cosh r - 1}{2}\left( 1 - \|a\|^2\right).
\]

\section*{Acknowledgement}
This research project was supported by the Japan Science and Technology Agency (JST), the Core Research of Evolutionary Science and Technology (CREST), the Center of Innovation Science and Technology based Radical Innovation and Entrepreneurship Program (COI Program), JSPS KAKENHI Grant No. JP 16H01707

\clearpage
\bibliographystyle{icml2020}
\bibliography{reference.bib}

\end{document}